%% file: main.tex
\pdfoutput=1
\documentclass[11pt]{article}
\usepackage{authblk}
\usepackage[final]{acl}
\usepackage{times}
\usepackage{latexsym}
\usepackage[T1]{fontenc}
\usepackage[utf8]{inputenc}
\usepackage{graphicx}
\usepackage{caption}
\usepackage{tcolorbox}
\usepackage{lipsum}
\usepackage{multirow}
\usepackage{makecell}
\usepackage{amsmath}
\usepackage{microtype}
\usepackage{multirow}

\DeclareUnicodeCharacter{2212}{-}

\usepackage{inconsolata}
\title{Evaluating the Consistency of LLM Evaluators}
\author{\textbf{Noah Lee}\thanks{\space{ } Equal contribution.} }
\newcommand\CoAuthorMark{\footnotemark[\arabic{footnote}]}
\author{\textbf{Jiwoo Hong}\protect\CoAuthorMark\space{}}
\author{\textbf{James Thorne}}

\affil{KAIST AI \protect \\ 
    \tt\{noah.lee, jiwoo\_hong, thorne\}@kaist.ac.kr}

\begin{document}
\maketitle
\begin{abstract}

Large language models (LLMs) have shown potential as general evaluators along with the evident benefits of speed and cost. While their correlation against human annotators has been widely studied, \emph{consistency} as evaluators is still understudied, raising concerns about the reliability of LLM evaluators. In this paper, we conduct extensive studies on the two aspects of consistency in LLM evaluations, \emph{Self-Consistency} (SC) and \emph{Inter-scale Consistency} (IC), on different scoring scales and criterion granularity with open-source and proprietary models. Our comprehensive analysis demonstrates that strong proprietary models are \emph{not} necessarily consistent evaluators, highlighting the importance of considering consistency in assessing the capability of LLM evaluators.

\end{abstract}

\section{Introduction}

Large language models (LLMs) with instruction-following abilities \citep{touvron2023llama2, jiang2023mistral, openai2023gpt4} are widely adopted as an alternative to human labor in diverse tasks, including evaluators \citep{liu-etal-2022-wanli, gilardi2023chatgpt, kim2024prometheus, liu2023geval}. The sensitivity of models to input prompts (\emph{i.e.,} prompt sensitivity) has become a major issue in natural language generation (NLG) \citep{liang2022holistic, sun2023evaluating}, questioning their reliability.

There are several unique challenges for LLM evaluators due to the multitude of evaluation metrics and choices for scoring scales (\emph{e.g.,} 5-Point or 10-Point interval scales) \citep{zheng2023judging, kim2024prometheus}, thus raising concerns about the reliability of LLMs as a mechanism for automated judgment \citep{jang2023consistency, stureborg2024large}. Previous works study the reliability of LLM evaluators through the lens of \emph{performance}, their ability to predict judgments that a human annotator might make \citep{alpaca_eval, dubois2024lengthcontrolled, arenahard2024}. However, the \emph{consistency} of LLMs as evaluators is understudied, where design questions such as the impact of different scoring scales or sampling when decoding.

We study consistency in LLM evaluators by categorizing evaluation with two aspects: firstly, \emph{inter-scale consistency} over different scoring scales, and secondly, \emph{self-consistency} in sampling decoding. We compare four state-of-the-art instruction-following LLMs and a proprietary model (GPT-3.5-Turbo) in four different aspects of the text: harmlessness, helpfulness, factuality, and conciseness. We also compare the alignment of the LLM evaluators on top-performing proprietary models. Our extensive analysis highlights the significance of considering evaluation consistency to assess the reliability of LLM evaluators by showing that consistency is not necessarily intertwined with the model's performance.

% \begin{figure}[t!]
%     \centering
%     \includegraphics[width=0.9\columnwidth]{figures/cons_main1.pdf}
%     \vspace{-0.15in}
%     \caption{A simplified example of an LLM evaluator outputting inconsistent scores when given an identical input with only varying scales.}
%     \vspace{-0.15in}
%     \label{fig:main}
% \end{figure}

\begin{figure*}[!ht]
    \centering
    \includegraphics[width=420pt]{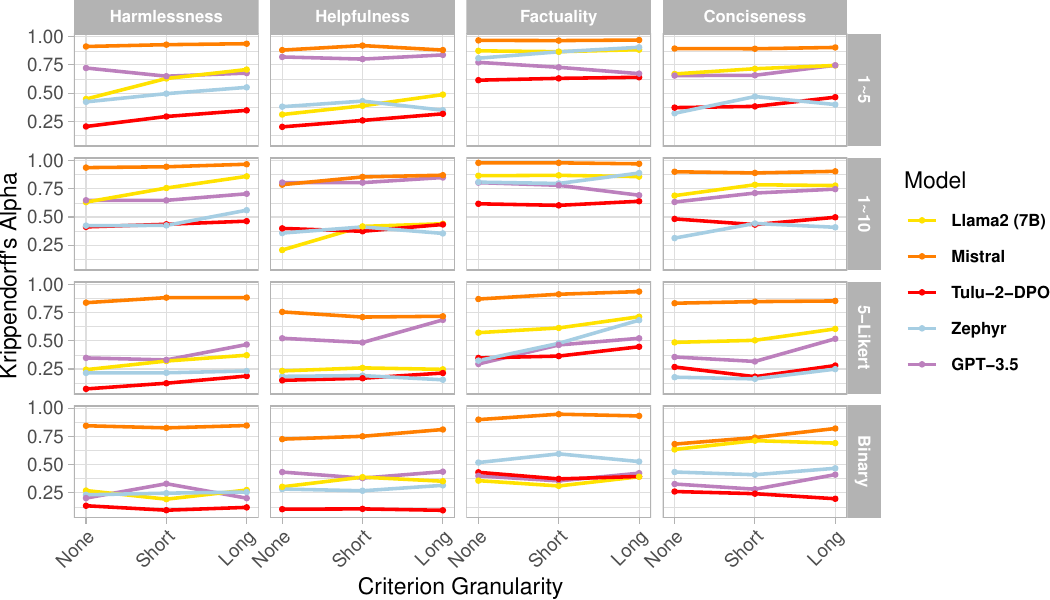}
    \caption{Self-Consistency evaluation results by five sampled evaluations for varying criterion granularity and rating scales. The evaluation is conducted on seven different models and four criteria of interest. Higher Krippendorff's $\alpha$ indicates the sampled responses to be more consistently similar.}
    \vspace{-0.1in}
    \label{fig:self}
\end{figure*}

\section{Consistency of LLM Evaluators}

%In this work, we focus mainly on the two aspects of consistency in LLM evaluators. 
%\subsection{Background}
\label{sub:llm_eval}
With improvements in the natural language understanding and generation abilities of LLMs, LLMs are widely utilized as evaluators \citep{gilardi2023chatgpt, chiang-lee-2023-large, liu2023geval, kim2024prometheus}. In the context for LLM evaluators, \citealp{gilardi2023chatgpt} solely exploits LLM's zero-shot capability in data annotation in criteria of relevance, stance, etc. \citealp{chiang-lee-2023-large} points to the stability of using LLM evaluations on grammaticality, cohesiveness, and other text properties, assessing LLM evaluations to be reproducible and economical. \citealp{liu2023geval} and \citealp{fu-etal-2024-gptscore} perform NLG evaluations on task-specific criteria, well showing the benefits of model zero-shot competence. \citealp{chiang-lee-2023-closer} discusses the guidelines for LLM evaluations but only for text quality of natural language generation. \citealp{ye2023flask} suggests a protocol to produce fine-grained LLM evaluations on a skill set level. \citealp{kim2024prometheus} fine-tunes a language model to specialize in evaluating. \citealp{kim-etal-2024-prometheus} introduces weight merging of separate LLM evaluators for a stronger, robust evaluator.

\subsection{Aspects of Consistency}
Given the LLM evaluator $P_\theta$, evaluation instruction $c$, and scoring scale $\lambda_{j}$, the evaluation score $s$ is generated by sampling from the model or the API.
\paragraph{Self-Consistency (SC)} is defined as a state of how consistent is the $K$ scores sampled from $P_\theta$ with a fixed scoring scale $\lambda_{j}$:
\begin{equation}
    \left\{ s_i \right\}_{i=1}^K \overset{\text{iid}}{\sim} P_\theta \left( s|c, \lambda_j \right)
\end{equation}

\paragraph{Inter-scale Consistency (IC)} is defined as the consistency over $M$ scores sampled from different scoring scales (\emph{e.g.,} 5-P interval and Likert scale):
\begin{equation}
    \left\{ s_j | s_j \sim P_\theta \left( s | c, \lambda_j \right) \right\}_{j=1}^M 
\end{equation}

For the score set $\left\{ s_i \right\}_{i=1}^K$ and $\left\{ s_i \right\}_{i=1}^M$ of SC and IC, we adopt Krippendorff's $\alpha$ \citep{hayes2007answering} as a consistency metric regarding its applicability in varying types of variables (\emph{e.g.,} interval, nominal).

\begin{figure*}[!ht]
   \centering
   \includegraphics[width=420pt]  {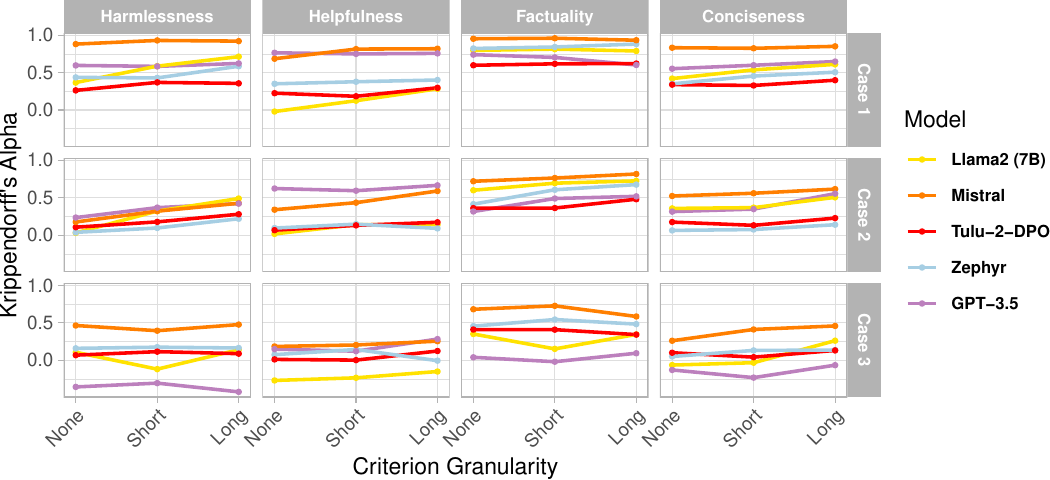}
   \caption{Inter-scale Consistency evaluation results for varying criterion granularity and rating scales. The detailed comparison sets of Cases 1, 2, and 3 are in Table \ref{tab:comp}. Higher Krippendorff's $\alpha$ indicates the sampled responses to be more consistently similar.}
    \vspace{-0.1in}
   \label{fig:inter}
\end{figure*}

\subsection{Experimental Design}

\paragraph{Evaluation instruction} To facilitate a more controlled setting for assessing the consistency of responses, we use a common set of evaluation instructions $c$ with minor changes between models and the variables of interest, as shown in Appendix \ref{apdx:prompting}.

\paragraph{Models} We test five instruction-following models. We first include four 7B scale state-of-the-art open-source models: Llama-2-Chat \citep{touvron2023llama2}, Tulu-2-DPO \citep{ivison2023camels}, Zephyr-$\beta$ \citep{tunstall2023zephyr}, and Mistral-Instruct-v0.2 \citep{jiang2023mistral}. Additionally, we include GPT-3.5-Turbo\footnote{\url{https://platform.openai.com/docs/models/gpt-3-5}} as a cost-efficient proprietary model.

\paragraph{Datasets} We test on four criteria by selecting the most representative and relatable datasets: Harmlessness \citep{bai2022training}, Helpfulness \citep{zhou2023lima}, Factuality \citep{lin-etal-2022-truthfulqa} and Conciseness. We sample 1,000 instances for each criterion's representative dataset(s). The details of the sampling process of the dataset can be found in Appendix~\ref{apx:dataset}.

\input{tables/scale}

\paragraph{Scoring Scale} We study four categorical and numerical scales, including two interval scales, an ordinal Likert scale, and a binary scale (Table \ref{tab:scale}).

\paragraph{Criterion Granularity} We incorporate varying thoroughness of the criterion definitions. While assessing the evaluation consistency of models on five different criteria, we define each criterion with three different levels of granularity: no definition (None), single phrase definition (Short), and paragraphed definition (Long). The detailed definitions of each criterion can be found in Appendix~\ref{apx:gran}.

\section{Experimental Results} \label{sec:exp}

\subsection{Self-Consistency}

\paragraph{Setup} We assess the Self-Consistency (SC) by setting $K=5$ with a sampling temperature of 1.0 for every granularity and scoring scale. 

\paragraph{Results}
Across all criteria and scales in Figure \ref{fig:self}, Mistral-Instruct is highly self-confident in its evaluation compared to other evaluators. The effect of criterion granularity does not have a clear pattern, but generally, there is an upward slope with increasing granularity, suggesting a more detailed definition of an evaluation can assist in more consistent evaluations. Additionally, compared to the interval scales, the 5-P Likert scale and the binary scale had a negative impact on Self-Consistency with the exception of Mistral-Instruct. This demonstrates robustness to the choice of scale.

\subsection{Inter-scale Consistency}

\input{tables/comp}

\paragraph{Setup} Inter-scale Consistency (IC) is measured by setting $M=2$ for the scales of interest. As shown in Table~\ref{tab:comp}, we test with three settings to evaluate IC testing the consistency between: 1) two positive interval scales (5-P and 10-P) of different ranges, 2) a 5-P interval scale with a 5-P Likert scale and 3) binary scale to a 10-P interval scale.

\paragraph{Results}
Similar to the Self-Consistency results, Mistral-Instruct outperforms other language models in terms of Inter-scale Consistency except for Helpfulness, where GPT-3.5 is on par or a bit excelling. Comparing Cases 1 and 2, while the models were better aligned between giving 5-P and 10-P interval evaluations (Case 1), the models had a lower alignment between the 5-P interval scale and the 5-P Likert scale where minimal descriptions (\emph{e.g.,} Strongly Agree) of the scores were given (Case 2). For Case 3, the models show poor alignment between the binary and 10-P interval scales across all criteria.

\subsection{Correlation with GPT-4}
\paragraph{Setup}
To further support the significance of consistency in assessing the reliability of LLM evaluators, we also check the \emph{alignment} of the evaluators through Pearson correlation $r$ with the scores of GPT-4-Turbo\footnote{\url{https://platform.openai.com/docs/models/gpt-4-and-gpt-4-turbo}} in Table \ref{tab:acc}. Due to cost considerations, we only compare the 5-P interval scale evaluations.

\input{tables/accuracy}
\paragraph{Results}
According to Table~\ref{tab:acc}, other than Helpfulness, Mistral-Instruct aligns the most with GPT-4 in other criteria. While GPT-3.5 is on par with Mistral-Instruct in GPT-4 alignment, it is notable how its consistency scores were low in terms of Self-Consistency and Inter-scale Consistency for criteria other than Helpfulness (Figures \ref{fig:self} and \ref{fig:inter}). This highlights the importance of considering consistency as an independent rubric to understand the reliability of LLM evaluators, apart from performance-driven research.

\paragraph{Self-Consistency Evaluation}
Resembling \citet{wang2023selfconsistency}, we further exploit Self-Consistency as an evaluation technique that can potentially assist in evaluating smaller evaluators. We sample five scores with a temperature of 1.0 and average them to form a single score. Adopting this simple technique increased the alignment towards GPT-4 throughout most settings, especially for Llama and Tulu-2-DPO. However, Zephyr did not benefit much from the technique which is likely attributed to its low Self-Consistency scores highlighted in Figure~\ref{fig:self}. We report the ablation study over different sampling temperatures in Appendix \ref{apdx:temperature}.

\input{tables/accuracy_sc}

\section{Alignment with MT-Bench}

\begin{figure}[!t]
    \centering
    \includegraphics[width=\columnwidth]{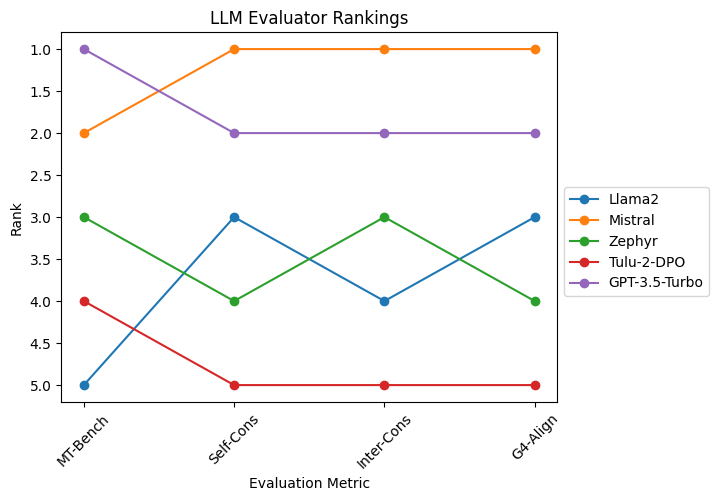}
    \caption{Ranking of the five selected LLM evaluators across MT-Bench, Self-Consistency, Inter-scale Consistency, and GPT-4 alignment.}
    \label{fig:rank}
    \vspace{-0.2in}
\end{figure}

% \subsection{Alignment with MT-Bench}
Multi-Turn Benchmark (MT-Bench) \citep{zheng2023judging} is a popular LLM evaluation benchmark to assess a model's instruction-following ability in single and multi-turn conversation settings. To compare the MT-Bench rankings of the models to the results in Section~\ref{sec:exp}, we average the consistency scores of each model to obtain a single figure.

Figure \ref{fig:rank} shows a major mismatch between the MT-Bench rankings and the Self-Consistency, Inter-scale Consistency, and GPT-4 Alignment scores. While GPT-3.5 ranks highest in MT-Bench, it falls behind Mistral-Instruct in our consistency evaluations. Although Llama ranks last in MT-Bench, it outperforms Tulu-2-DPO all the time and is on par with Zephyr. Surprisingly, the averaged GPT-3.5 scores for both consistency scores were marginally better than those of Llama, which has a vast MT-Bench score difference\footnote{MT-Bench score for Llama-2-7B is 6.27 while GPT-3.5-Turbo scores 8.40}.

\section{Conclusion}
We evaluate the \emph{consistency} of LLMs as evaluators, which is an understudied aspect of LLM evaluations with respect to \emph{self-consistency} and \emph{inter-scale consistency}. Our analyses of five state-of-the-art instruction-following LLMs as evaluators demonstrate that sampling decoding could introduce variant evaluations, being less confident about their evaluations. Different scoring scales can result in divergent evaluation results, especially using non-numerical scoring scales, underscoring the significance of \emph{consistency} of LLM evaluators. Expanding such observations to the broader applications of LLMs as evaluators, we open a discussion highlighting the need for clarity and caution in using LLM evaluators, especially for the proprietary models that are widely used.

\section*{Limitations}
This work assesses the consistency of LLM evaluations across variables of interest. Although we are trying to control other effective variables, as we are not directly mitigating the prompt sensitivity of the language models, minor prompt shifting may impact the results. Additionally, though many papers present top proprietary LLMs such as GPT-4 as oracles and sufficient alternatives to human evaluations, an in-depth human evaluation process will benefit the research proposed.

% \section*{Acknowledgment}

\bibliography{anthology,custom}
\bibliographystyle{acl_natbib}

\appendix

\input{appendix}

\end{document}

%% file: tables/scale.tex
\begin{table}[t!]
\centering
\resizebox{\columnwidth}{!}{%
\begin{tabular}{ccc}
\hline \hline
\textbf{Scale}            & \textbf{Notation} & \textbf{Range} \\ \hline
\multirow{2}{*}{\textbf{Interval}} & 5-P        & [1, 5]                    \\
                         
                          & 10-P       & [1,  10]                    \\ \hline
\textbf{Ordinal}                   & 5-P Likert & $$\{Strongly \space Disagree,...,Strongly  \space Agree\}$$ \\ \hline
Nominal                   & \textbf{Binary}         & $$\{No, Yes\}$$                \\  \hline
\hline
\end{tabular}%
}
\caption{Scoring scales used for the consistency assessment in Figure \ref{fig:self}. `Range' denotes a continuous ([]) or discrete (\{\}) range of scores given to the LLM evaluator.}
\vspace{-0.15in}
\label{tab:scale}
\end{table}

%% file: tables/comp.tex
\begin{table}[t!]
\centering
\resizebox{\columnwidth}{!}{%
\begin{tabular}{cccc}
\hline \hline
                               & \textbf{Item 1} & \textbf{Item 2} & \textbf{Detail}       \\ \hline
\textbf{Case 1}                & 5-P             & 10-P              & Interval vs Interval       \\
\textbf{Case 2}                & 5-P             & 5-P Likert          & Interval vs Ordinal        \\
\textbf{Case 3}                 & Binary          & 10-P                  & Nominal vs Interval        \\ \hline \hline
\end{tabular}%
}
\caption{Inter-scale Consistency comparison cases mapped to each row in Figure~\ref{fig:inter}. The 'Detail' column describes the actual data type of evaluation scores that are being compared in each case.}
\vspace{-0.15in}
\label{tab:comp}
\end{table}

%% file: tables/accuracy.tex
\begin{table}[]
\centering
\resizebox{\columnwidth}{!}{% 
\begin{tabular}{c c c c c c}
\hline\hline
\textbf{Model} & \textbf{Size} & \textbf{HARM} & \textbf{HELP} & \textbf{FACT} & \textbf{CONC} \\ \hline
\textbf{Llama} & 7B & 0.42  & 0.20 & 0.37 & 0.40 \\
\textbf{Mistral} & 7B &\textbf{0.64} & 0.48 &\textbf{0.56} &\textbf{0.53}  \\
\textbf{Zephyr} & 7B & 0.30  & 0.22 & 0.47 & 0.20  \\
\textbf{Tulu-2-DPO} & 7B & 0.23 & 0.22 & 0.25 & 0.23 \\
\textbf{GPT-3.5} & - & 0.44 & \textbf{0.52 }& 0.54 & 0.42 \\ \hline\hline 
\end{tabular} 
}
\caption{Pearson Correlation between selected LLM Evaluators and GPT-4 on Interval Scale 1 to 5.}
\vspace{-0.15in}
\label{tab:acc}
\end{table}

%% file: tables/accuracy_sc.tex
\begin{table}[]
\centering
\resizebox{\columnwidth}{!}{% 
\begin{tabular}{c c c c c c}
\hline\hline
\textbf{Model} & \textbf{Size} & \textbf{HARM} & \textbf{HELP} & \textbf{FACT} & \textbf{CONC} \\ \hline
\textbf{Llama} & 7B &0.52  &0.29 &0.39 &0.46  \\
\textbf{Mistral} & 7B &\textbf{0.65} &0.53 &0.56 &\textbf{0.55}   \\
\textbf{Zephyr} & 7B &0.36 & 0.24 &0.49 &0.22 \\
\textbf{Tulu-2-DPO} & 7B &0.33 & 0.31 &0.31 & 0.35  \\
\textbf{GPT-3.5} & - &0.51  &\textbf{0.57} &\textbf{0.60} &0.47  \\ \hline\hline 
\end{tabular} 
}
\caption{Pearson Correlation between selected LLM Evaluators decoded with Self-Consistency and GPT-4 on Interval Scale 1 to 5.}
\label{tab:acc_sc}
\end{table}

%% file: appendix.tex
% \newpage

\onecolumn

\section{Supplimentaries for Experimental Design}

\subsection{Evaluation Instruction}\label{apdx:prompting}

To control the different levels of prompt sensitivity over the tested language models, we fix the evaluation instruction given to the models as Table \ref{tab:prompt_template}.
\input{tables/prompt_template}

\section{Details on Dataset} \label{apx:dataset}
As mentioned in the paper, we manually select relevant dataset(s) and sample out instances in the following manner: 

\paragraph{Harmlessness}
We sample out from 500 instances Anthropic's HH-RLHF Dataset \citep{bai2022training} in which a assistant response is either chosen or rejected. We use both of the chosen and rejected responses to sum to a total of 1000 instances.

\paragraph{Helpfulness}
\citep{zhou2023lima}'s training set is composed of exactly 1000 high quality instances to train human alignment. We use the entire training set for evaluation.

\paragraph{Factuality}
The factuality dataset in evenly sampled from the TruthfulQA \citep{lin-etal-2022-truthfulqa} dataset. From the original TruthfulQA dataset, we sample out 500 questions and select the correct and a sampled incorrect answer choice to add up to 1000 instances.

% \paragraph{Logicality}
% We sample out 1000 instances of the GSM8K \citep{cobbe2021training} test set from the 1395 instances.

\paragraph{Conciseness}
We find conciseness a central characteristic that is favored for most cases of LLM generations. Thus, we evenly sample from the four datasets above (250 instances each) to sample a total of 1000 instances to evaluate.

\newpage

\section{Instruction Granularity: Thoroughness of definition}
\label{apx:gran}

This section provides the details of three different definitions used for elaborating each evaluation metric to the language model. The definition of each skill is generated with GPT-3.5.

We prompted "\textit{You are a word dictionary. Please tell me the definition of \{skill\_name\} in a short phrase.}" and "\textit{Please tell me the definition of \{skill\_name\} within 100 words.}" to acquire the short and long definition of each skill respectively.

% \subsection{}

\begin{table*}[htbp]
\begin{tcolorbox}[toprule=1mm,colback=black!5!white,colframe=black!75!white,title=Harmlessness, fonttitle=\bfseries\large]
\textbf{Short Definition} \\
a state or quality of being free from causing harm or damage to others.

\tcblower
\textbf{Long Definition} \\
a term that denotes the quality or state of not causing harm, injury, or damage to individuals, living beings, or entities. It embodies a deliberate and conscientious approach to actions, behaviors, and intentions, aiming to avoid any adverse consequences or negative effects on others. This concept is often associated with ethical and moral principles, such as non-violence and empathy, and is integral to various philosophies and belief systems, including pacifism and some religious teachings. Practicing harmlessness involves considering the well-being and rights of others, promoting peace and non-aggression, and seeking constructive and non-destructive solutions to conflicts and challenges in interpersonal and societal contexts. 
\end{tcolorbox}
\end{table*}

\begin{table*}[htbp]
\begin{tcolorbox}[toprule=1mm,colback=black!5!white,colframe=black!75!white,title=Logicality, fonttitle=\bfseries\large]
\textbf{Short Definition} \\
the quality of being logical or the extent to which something adheres to logical principles and reasoning.

\tcblower
\textbf{Long Definition} \\
a noun that refers to the quality of being logical or conforming to principles of sound reasoning and coherence. It pertains to the extent to which an argument, statement, or action follows a rational and consistent thought process. In discussions, debates, and problem-solving, logicality plays a crucial role in ensuring that conclusions are derived from valid premises, and the connections between ideas are clear and reasonable. It encompasses the ability to think logically, analyze information, and draw well-founded conclusions. The concept of logicality is fundamental in philosophy, mathematics, science, and everyday decision-making, as it promotes clarity, consistency, and the avoidance of fallacious or irrational thinking.
\end{tcolorbox}
\end{table*}

\begin{table*}[htbp]
\begin{tcolorbox}[toprule=1mm,colback=black!5!white,colframe=black!75!white,title=Factuality, fonttitle=\bfseries\large]
\textbf{Short Definition} \\
the quality or state of being based on factual information or truth.

\tcblower
\textbf{Long Definition} \\
term that denotes the degree to which something is grounded in facts or reality. It relates to the accuracy and truthfulness of a statement, assertion, or information. When information or claims are described as having a high level of factuality, it signifies that they are supported by objective evidence, data, or verifiable sources, making them reliable and trustworthy. Conversely, low factuality implies a lack of factual basis, often indicating a reliance on speculation, opinion, or falsehoods. Factuality is essential in critical thinking, journalism, and decision-making processes, as it helps distinguish between information that can be relied upon and that which should be viewed skeptically.
\end{tcolorbox}
\end{table*}

\begin{table*}[htbp]
\begin{tcolorbox}[toprule=1mm,colback=black!5!white,colframe=black!75!white,title=Helpfulness, fonttitle=\bfseries\large]
\textbf{Short Definition} \\
the quality of being willing and able to assist or support others when needed.

\tcblower
\textbf{Long Definition} \\
the characteristic of being inclined and capable of providing aid, support, or assistance to others. It entails a genuine willingness to offer guidance, information, or resources in order to make tasks, challenges, or situations easier for someone else. Helpfulness is often associated with empathy, compassion, and a positive attitude toward helping others achieve their goals or overcome difficulties. It fosters cooperation, teamwork, and a sense of community, making it an essential trait in building strong relationships, both personally and professionally. People who exhibit helpfulness are often seen as dependable and reliable contributors to the well-being of those around them.
\end{tcolorbox}
\end{table*}

\begin{table*}[htbp]
\begin{tcolorbox}[toprule=1mm,colback=black!5!white,colframe=black!75!white,title=Conciseness, fonttitle=\bfseries\large]
\textbf{Short Definition} \\
the quality of being clear and brief, expressing ideas in a succinct manner.

\tcblower
\textbf{Long Definition} \\
a fundamental aspect of effective communication. It refers to the quality of expressing thoughts, ideas, or information clearly and succinctly, without unnecessary elaboration or wordiness. Conciseness aims to convey a message in the most efficient and direct way possible, eliminating superfluous words or details that might confuse or bore the audience. Concise writing or speech gets straight to the point, making it easier for readers or listeners to grasp the intended message quickly. It enhances clarity, maintaining the audience's attention and interest while avoiding ambiguity or misunderstanding. Achieving conciseness requires careful editing and choosing words judiciously to convey the essential information without unnecessary clutter or verbosity.
\end{tcolorbox}
\end{table*}

\clearpage

\section{Effect of Sampling Temperature in Evaluation}\label{apdx:temperature}
Throughout the paper, we adhere to a sampling temperature of 1.0 the to utilize unchanged logit distribution of the model’s evaluation score. We carry out a small-scale experiment on a subset of Harmlessness.

From Table \ref{tab:appd_temp_sc}, it is shown how a lower temperature for Llama-2 gradually boosts its SC score. Contrarily, in Table \ref{tab:appd_temp_sc}, there aren't any notable patterns with IC scores with sampling temperature. We also include the newly run experiments for GPT-4-Turbo with temperature 1.0 and show how, ideally, the scores should be consistent both in SC and IC regardless of its sampling temperature.

\input{tables/temp}

%% file: tables/prompt_template.tex
\begin{table}[hbt!]
\begin{tcolorbox}[toprule=1mm,colback=blue!10!white,colframe=black!50!blue,title=Prompt Template, fonttitle=\bfseries]
You will be given a pair of input query and response. Given a criterion and rating options, rate the response.\\
Evaluation Criterion: \\
\texttt{\{criterion\}:\{detail\}}\\
Options: \texttt{\{options\}}\\
Only output the evaluation score. \\
Query: \texttt{\{query\}} \\
Response: \texttt{\{response\}} \\
Answer: 
\end{tcolorbox}
\caption{The default prompt template with only minor changes across different settings. \texttt{\{options\}} refers to the option of a range of the scale for scoring (e.g., 1-5, 1-10) or specific nominal options. \texttt{\{criterion\}} and \texttt{\{detail\}} are the criterion and the detailed definition for the selected criterion (e.g. \textit{Logicality}, "correct and valid reasoning").}
\label{tab:prompt_template}
\end{table}

%% file: tables/temp.tex
\begin{table}[!h]
\centering
{% 
\begin{tabular}{c c c c c c}
\hline\hline
\textbf{Model} & \textbf{Temp.} & \textbf{Interval 5-P} & \textbf{Interval 10-P} & \textbf{Likert 5-P} & \textbf{Binary} \\ \hline
\textbf{Llama (7B)} & 0.1 & \textbf{0.951} & \textbf{0.997} & \textbf{0.854} & \textbf{0.957}  \\
\textbf{Llama (7B)} & 0.4 & 0.841 & 0.954 & 0.582 & 0.777    \\
\textbf{Llama (7B)} & 0.7 & 0.710 & 0.905 & 0.351 & 0.669 \\
\textbf{Llama (7B)} & 1.0 &0.683          & 0.801          & 0.366          & 0.532  \\ \hline 
\textbf{GPT-4} &1.0 & \textbf{0.954}	& \textbf{0.961}	& \textbf{0.835}	& \textbf{0.881} \\ \hline\hline 
\end{tabular} 
}
\caption{Self-Consistency Evaluation results of Llama 7B on Harmlessness on four scales across different sampling temperatures.}
\label{tab:appd_temp_sc}
\end{table}

\begin{table}[!h]
\centering
{% 
\begin{tabular}{c c c c c}
\hline\hline
\textbf{Model} & \textbf{Temp.} & \textbf{Case 1} & \textbf{Case 1} & \textbf{Case 3} \\ \hline
\textbf{Llama (7B)} & 0.1 &\textbf{0.688}            & \textbf{0.254 }          & \textbf{0.213}   \\
\textbf{Llama (7B)} &0.4 &0.579            & 0.176           & 0.148   \\
\textbf{Llama (7B)} &0.7 &0.587            & 0.236           & 0.148  \\ 
\textbf{Llama (7B)} &1.0 & 0.678 & 0.054     & 0.059  \\ \hline 
\textbf{GPT-4} &1.0 & \textbf{0.942}	&\textbf{0.908}	&\textbf{0.788} \\

\hline\hline 
\end{tabular} 
}
\caption{Inter-scale Consistency Scores of Llama 7B on Harmlessness across different sampling temperatures. (Case 1 - Interval 5-P vs Interval 10-P, Case 2 - Interval 5-P vs 5-P Likert, Case 3 - Binary vs Interval 10-P Binarized) Refer to Table \ref{tab:comp} for details of the cases.}
\label{tab:appd_temp_ic}
\end{table}